\documentclass[manuscript,screen,nonacm]{acmart} 
\setcopyright{cc} 
\acmConference[]{} 

\newcommand{\workshopname}{GenAICHI: CHI 2025 Workshop on Generative AI and HCI}
\newcommand\extrafootertext[1]{
    \bgroup
    \renewcommand\thefootnote{\fnsymbol{footnote}}%
    \renewcommand\thempfootnote{\fnsymbol{mpfootnote}}%
    \footnotetext[0]{#1}%
    \egroup
}

\AtBeginDocument{ 
    \fancypagestyle{firstpagestyle}{
        \fancyhf{}
        \fancyfoot[L]{\sffamily\footnotesize \workshopname}%
        \fancyfoot[C]{\sffamily\footnotesize \thepage}
    }
    \fancyhf{}
    \fancyhead[L]{\sffamily\footnotesize\shorttitle}
    \fancyhead[R]{\sffamily\footnotesize\shortauthors}
    \fancyfoot[L]{\sffamily\footnotesize\workshopname}%
    \fancyfoot[C]{\sffamily\footnotesize\thepage}
}

\AtBeginDocument{%
  }

\setcopyright{acmlicensed}
\copyrightyear{2025}
\acmYear{2025}
\acmConference[CHI '25]{GenAICHI: CHI 2025 Workshop on Generative AI and HCI}{June 03--05, 2018}{Woodstock, NY}

\begin{document}

\title{Making Physical Objects with Generative AI and Robotic Assembly: Considering Fabrication Constraints, Sustainability, Time, Functionality and Accessibility}


\author{Alexander Htet Kyaw}
\email{alexkyaw@mit.edu}
\orcid{0000-0002-6020-4529}
\affiliation{%
  \institution{Massachusetts Institute of Technology}
    \country{United States}
}

\author{Se Hwan Jeon}
\email{sehwan@mit.edu}
\affiliation{%
  \institution{Massachusetts Institute of Technology}
    \country{United States}
}

\author{Miana Smith}
\email{miana@mit.edu}
\affiliation{%
  \institution{Massachusetts Institute of Technology}
    \country{United States}
}

\author{Neil Gershenfeld}
\email{neil.gershenfeld@cba.mit.edu}
\affiliation{%
  \institution{Massachusetts Institute of Technology}
    \country{United States}
}

\renewcommand{\shortauthors}{Kyaw et al.}

\settopmatter{printacmref=false} 
\renewcommand\footnotetextcopyrightpermission[1]{} 
\makeatletter 
\renewcommand{\@authorsaddresses}{Licensed under a Creative Commons Attribution 4.0 International License (CC BY 4.0). Copyright remains with the author(s).} 
\makeatother 

\begin{abstract}
3D generative AI enables rapid and accessible creation of 3D models from text or image inputs. However, translating these outputs into physical objects remains a challenge due to the constraints in the physical world. Recent studies have focused on improving the capabilities of 3D generative AI to produce fabricable outputs, with 3D printing as the main fabrication method. However, this workshop paper calls for a broader perspective by considering how fabrication methods align with the capabilities of 3D generative AI. As a case study, we present a novel system using discrete robotic assembly and 3D generative AI to make physical objects. Through this work, we identified five key aspects to consider in a physical making process based on the capabilities of 3D generative AI. 1) Fabrication Constraints: Current text-to-3D models can generate a wide range of 3D designs, requiring fabrication methods that can adapt to the variability of generative AI outputs. 2) Time: While generative AI can generate 3D models in seconds, fabricating physical objects can take hours or even days. Faster production could enable a closer iterative design loop between humans and AI in the making process. 3) Sustainability: Although text-to-3D models can generate thousands of models in the digital world, extending this capability to the real world would be resource-intensive, unsustainable and irresponsible. 4) Functionality: Unlike digital outputs from 3D generative AI models, the fabrication method plays a crucial role in the usability of physical objects. 5) Accessibility: While generative AI simplifies 3D model creation, the need for fabrication equipment can limit participation, making AI-assisted creation less inclusive. Discrete robotic assembly with 3D generative AI shows promising results in addressing key aspects such as fabrication constraints, time, sustainability, and functionality, though accessibility remains a limitation. These five key aspects provide a framework for assessing how well a physical making process aligns with the capabilities of 3D generative AI and values in the world. 

\end{abstract}





\maketitle

\section{Introduction}
Recent advances in 3D generative AI have enabled the fast and easy creation of digital models from text descriptions or image inputs \cite{gozalo-brizuela_survey_2023}. For example, models like InstantMesh, Latte3D, GET3D, and Shap-E allow users to generate 3D models in seconds  \cite{xu_instantmesh_2024} \cite{xie_latte3d_2024} \cite{gao_get3d_2022} \cite{jun_shap-e_2023}. 
However, while AI-generated 3D models are easy to produce, fabricating them into physical objects remains a challenge due to real-world constraints \cite{faruqi_style2fab_2023} \cite{makatura_how_2023}. Previous research has focused on improving capabilities of generative AI to produce fabricable outputs, with 3D printing as the primary fabrication method \cite{edwards_sketch2prototype_2024} \cite{danry_organs_2023} \cite{faruqi_shaping_2024} \cite{mcclelland_generative_2022}. However, 3D printing as a fabrication method has significant limitations \cite{yildirim_digital_2020}. While it allows for precise manufacturing, it can take hours or even days to produce a single object, making real-time human-AI collaboration more challenging \cite{kim_understanding_2017} \cite{zhou_understanding_2024} \cite{dogan_fabricate_2022}. Additionally, desktop 3D printing is best suited for small to medium-sized and tabletop-scale objects, often constrained by the size of the printer bed \cite{chen_dapper_2015} \cite{prusa_original_nodate}. This limitation prevents the fabrication of larger structures, such as furniture or room-scale objects. As generative AI continues to advance, fabrication methods should algin with its speed and scale, enabling on-demand, large-scale production in a sustainable manner. Relying on 3D printing or only improving the capabilities of 3D generative AI models is insufficient to fully realize the potential of generative AI in physical making. This paper calls for a broader perspective by considering how the making process aligns with the capabilities of generative AI. 

\begin{figure}
    \centering
    \includegraphics[width=1\linewidth]{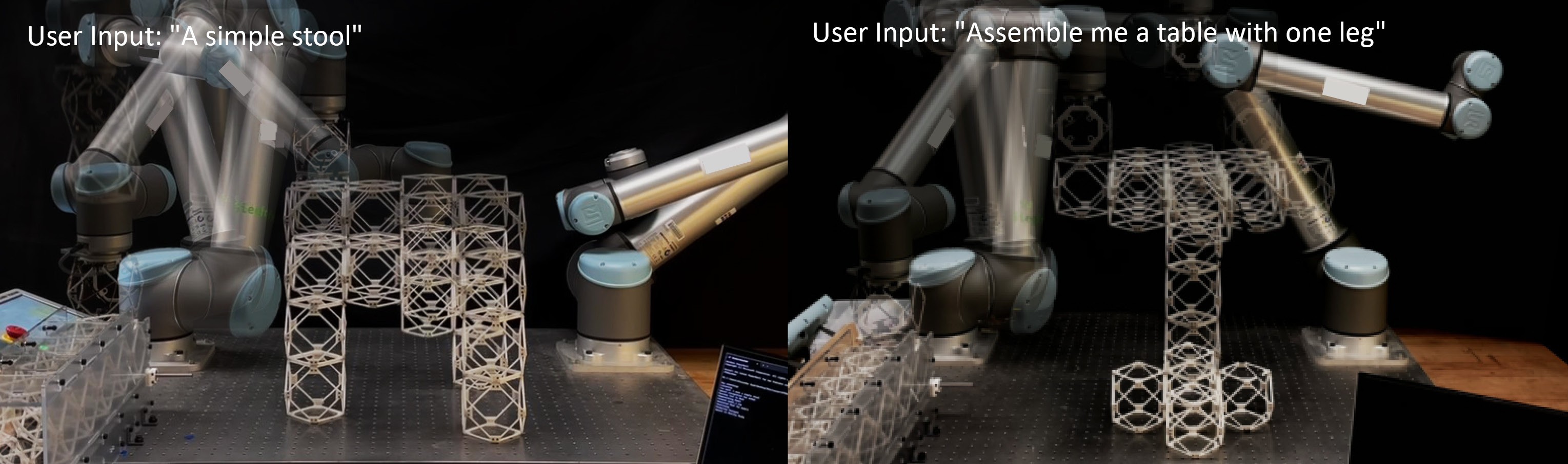}
    \caption{Figure 1. Demonstration of 3D Generative AI to Robotic Assembly from a User Prompt.
}
    \Description{A robotic arm assembling structures based on natural language user input. On the left, the robot constructs a simple stool in response to the prompt “A simple stool.” On the right, the robot assembles a table with one leg following the command “Assemble me a table with one leg.” The motion blur effect illustrates the robotic arm’s movement during the construction process. The framework structures are composed of modular lattice components.}
    \label{fig:1}
\end{figure}

To address this, we explore discrete robotic assembly as an alternative fabrication process for 3D generative AI outputs \cite{kyaw_speech_2024}. By leveraging generative AI, our approach lowers the barrier to entry for robotic assembly, making it possible for users to speak objects into reality without programming or 3D modeling \cite{ballagas_exploring_2019}. Through this case study, we propose a framework for evaluating fabrication methods based on five key aspects: fabrication constraints, production time, sustainability, functionality, and accessibility. The workshop paper contributes to bridging the gap between generative AI and physical making by examining how fabrication processes can better support AI-generated designs. We propose a novel system using 3D generative AI and discrete robotic assembly as an on-demand, prompt-to-production method with the goal of creating functional objects in a sustainable and accessible manner. (Fig \ref{fig:1})
\section{Method}

\begin{figure}
    \centering
    \includegraphics[width=1\linewidth]{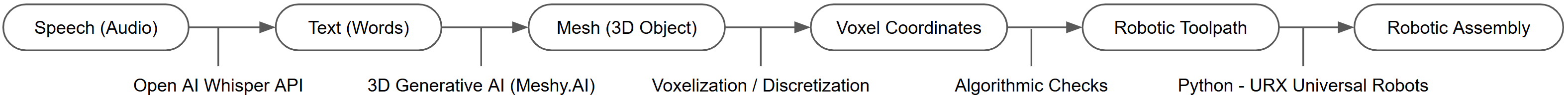}
    \caption{System pipeline from speech to robotic assembly}
     \Description{A sequential flowchart representing a pipeline from speech input to robotic assembly. The process begins with OpenAI Whisper API converting speech to text. The text is then processed by 3D generative AI (Meshy.AI) to create a 3D mesh object. This mesh is discretized into voxel coordinates, undergoing algorithmic checks before generating a robotic toolpath. The toolpath is executed via Python using URX for Universal Robots, enabling robotic assembly.}
    \label{fig:2}
\end{figure}

In order to convert speech into physical objects, we present a system that uses discrete robotic assembly and 3D generative AI, employing a 6-axis robotic arm to assemble smaller modular components into larger objects. In our system, we use the UR10 robot from Universal Robots and Meshy.AI as the text-to-3D generative AI model \cite{universal_robots_ur10e_2025}\cite{meshy_meshy_2025} (Fig \ref{fig:2}). Each paragraph investigates 3D generative AI-based discrete robotic assembly through five key aspects: fabrication constraints, production time, sustainability, functionality, and accessibility.

Fabrication Constraints: 3D generative AI models typically output meshes or point clouds \cite{li_generative_2024} \cite{poole_dreamfusion_2022}. However, they do not inherently consider fabrication constraints or generate component-level representations suitable for robotic assembly. Additionally, AI models can generate a wide variety of 3D geometries, requiring the making process to adapt to the variability of generative AI outputs. To address this, we discretize the AI-generated geometry into assembly components through a voxelization algorithm based on the dimensions of the assembly component. While this provides a component-level representation for robotic assembly, it does not guarantee that the robot will successfully complete the assembly. To ensure assembly feasibility, we further modify the geometry based on algorithmic checks and fabrication constraints, including overhang detection, connectivity search, and robotic arm reachability analysis.

Time: While generative AI can produce 3D models in seconds, manufacturing physical objects takes considerably longer. Faster assembly could enable a closer iterative design loop between humans and AI in the making process in the future \cite{gmeiner_exploring_2023}. Unlike traditional manufacturing methods, discrete robotic assembly relies on prefabricated modular parts, reducing the need for material preprocessing. However, we acknowledge that other methods, such as 3D printing, also require material preparation (e.g., filament spooling, resin curing). For a fair comparison, we measure fabrication time from the user's request, excluding material preparation and prefabrication. 

Sustainability:  3D Generative AI enables the creation of countless assets, but translating this capability into the physical world can be resource-intensive, unsustainable, and irresponsible \cite{savage_objectify_2023}. We study how discrete robotic assembly can mitigate material waste by enabling the disassembly and reassembly of components into different objects. We start with 40 modular components and evaluate how many distinct objects can be assembled from the same set of parts. The goal of this method is to promote sustainable fabrication by ensuring that physical artifacts generated through AI can be efficiently repurposed rather than discarded after a single use.

Functionality: As text-to-3D models advance, it is also crucial from the fabrication process to be able to make objects that are functional. For example, if the AI generates a chair, the robotic system may successfully assemble a chair-like structure. However, if this object lacks structural integrity and collapses when someone sits on it, it could pose a safety risk. In our case study, we assemble lattice-based cuboidal components with magnets on each face for secure yet reversible connections \cite{jenett_materialrobot_2019}. Our goal is to create functional objects that can withstand some amount of compression.

Accessibility: While generative AI offers an accessible method for creating 3D models, physical fabrication often requires specialized equipment and hardware. This system can be implemented using an industrial robotic arm, a low-cost robotic arm, or DIY robotic arms assembled at home from 3D-printed parts. To assess accessibility, we calculate the cost of the industrial robotic arm and alternatives. Expanding accessibility through lower-affordable options can make AI-driven fabrication more inclusive, enabling broader participation \cite{kuznetsov_rise_2010}.

\section{Results}

\begin{figure}
    \centering
    \includegraphics[width=1\linewidth]{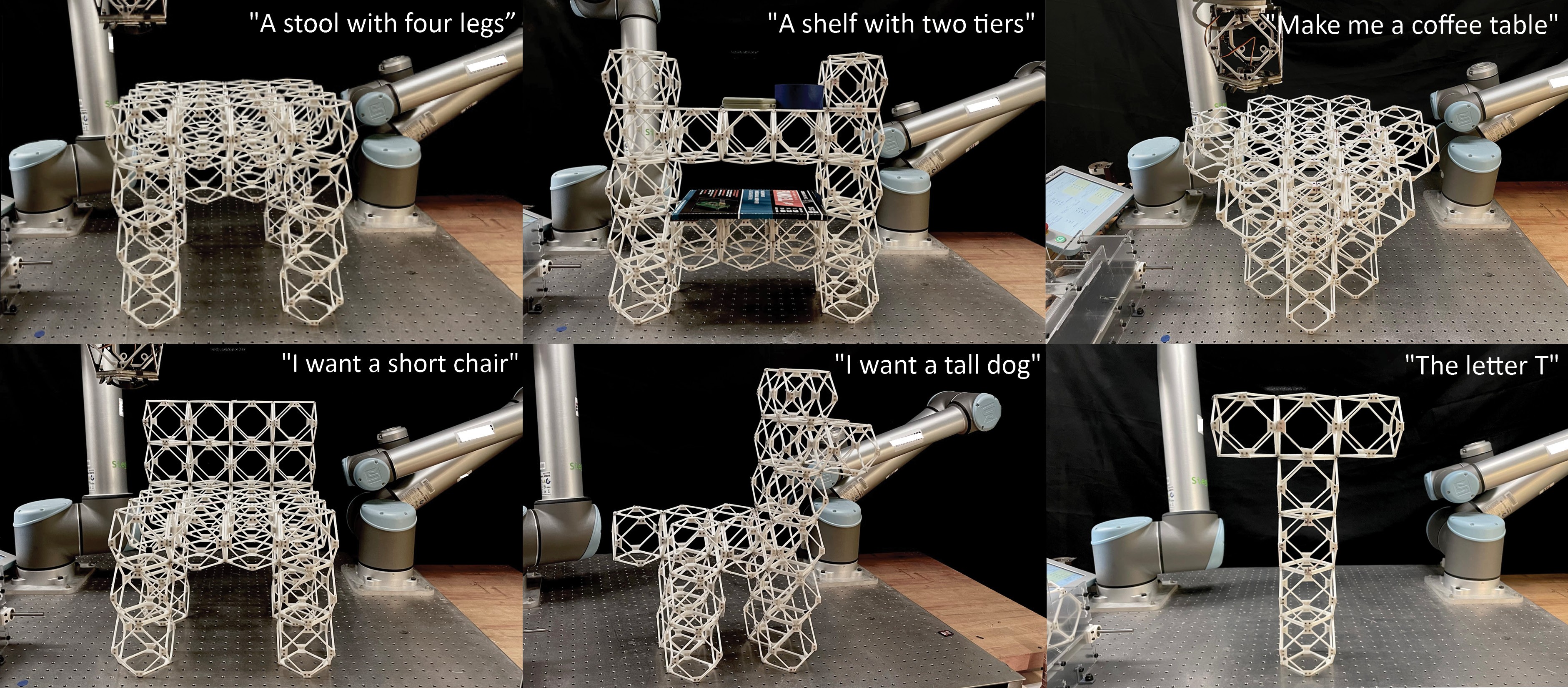}
    \caption{A variety of objects created from user prompts using 3D generative AI and discrete robotic assembly}
    \Description{A set of six images showing a robotic arm assembling modular lattice structures based on natural language commands. The images display different objects constructed by the robot, including a four-legged stool, a two-tier shelf, a coffee table, a short chair, a tall dog-like structure, and a letter "T." Each object is built using interconnected lattice components}
    \label{fig:3}
\end{figure}

Our results show that discretizing AI-generated meshes into component-level representations can handle the variability in generative AI outputs for discrete robotic assembly. (Fig \ref{fig:3}) However, simply voxelizing an AI-generated mesh into a component-level representation does not always guarantee successful robotic assembly. Our experiments showed that most attempts failed without overhang detection, connectivity analysis, and robotic arm reachability assessment. 

For assembly time, we measured the total duration required to assemble each object, starting from the moment the user prompts the system. The assembled objects have an average volume of around  7500 cm³, with nearly all of them being assembled in under five minutes. Considering that generative AI can produce objects of any size in seconds, we believe that the ability of discrete robotic assembly to construct large objects within minutes presents a promising match. With a fast prompt-to-physical workflow, future research could explore an iterative design loop between humans and generative AI in the making process.

Our results show that discrete robotic assembly enables sustainable production through the reuse of modular components to construct different objects. Using a set of 40 modular components, we successfully assembled seven distinct objects from the same set. This finding highlights the potential for material-conscious and responsible physical making with generative AI. Unlike traditional manufacturing methods, where material waste accumulates after each fabrication cycle, discrete robotic assembly enables disassembly and repurposing. 

To assess functionality, we check both the shape and usability of the object. We use GPT-4’s vision-language capabilities to verify whether the assembled object corresponds to the user prompt \cite{openai_openai_2025} . Our experiments with the vision language model indicate that all assembled objects resemble the user prompt. As for the usability of the objects, the chair and stool we assembled could not support sitting, whereas the shelf and table functioned as intended, allowing us to place objects on them. The Letter T and the tall dog were evaluated only based on their visual resemblance using GPT-4.

Our system enables users to create physical objects from speech without requiring expertise in 3D modeling or robotic programming. However, cost is the main accessibility barrier of this system. An industrial UR10 robotic arm used in this study can cost between \$40,000 and \$50,000 \cite{qviro_ur10e_2025} . DIY robotic arms constructed from 3D-printed parts or cheaper off-the-shelf robotic arms can cost under \$1,000 but have reduced precision and payload capacity \cite{robotics_elephant_2025} \cite{jenett_materialrobot_2019}. However, further tests with low-cost or 3D-printed robotic arms are needed, as this study was conducted with a UR10 robotic arm.

\section{Conclusion}
 While current text-to-3D outputs cannot be directly assembled, we demonstrate that discretizing them into component-level representations can accommodate variability in AI-generated designs and enable robotic assembly. We show that overhang detection, connectivity search, and robotic arm reachability analysis are necessary since 3D generative AI outputs do not inherently consider physical constraints. The rapid creation of physical objects aligns with AI’s fast generative capabilities, opening new research avenues for future studies on real-time, human-AI iterative design workflows in physical fabrication. Future projects can also explore the user of augmented reality to enhance human-machine collaborating with robots during the assembly process \cite{kyaw_gesture_2024} \cite{kyaw_humanmachine_2024}

 Furthermore, reusing components enables a more sustainable and responsible approach to making with generative AI outputs compared to wasteful manufacturing methods. By leveraging generative AI, our approach contributes to lowering the barrier to robotic assembly, enabling users to speak objects into reality. However, challenges remain, particularly in ensuring the functionality of certain objects and accessibility. Future work could enhance the structural integrity and functionality of objects while exploring the integration of generative AI with discrete robotic assembly across varying component types, and connection methods. 
 
 This workshop paper contributes to a broader framework on how fabrication methods align with the capabilities of 3D generative AI. Additionally, the study contributes to the advancement of AI-driven fabrication by integrating 3D generative AI with discrete robotic assembly, enabling on-demand, sustainable, and accessible production.


\bibliographystyle{ACM-Reference-Format} \bibliography{references}

\end{document}